\documentclass[letterpaper]{article} 
\usepackage{mysty}  
\usepackage{times}  
\usepackage{helvet}  
\usepackage{courier}  
\usepackage[hyphens]{url}  
\usepackage{graphicx} 
\usepackage{natbib}  
\usepackage{caption} 
\usepackage{algorithm}
\usepackage{algorithmic}
\usepackage{bm}
\usepackage{amsfonts}
\usepackage{multirow}
\usepackage{booktabs} 
\usepackage{amssymb}
\usepackage{mathrsfs}
\usepackage{color}

\usepackage{hyperref}
\hypersetup{hypertex=true,
            colorlinks=true,
            linkcolor=blue,
            anchorcolor=blue,
            citecolor=blue}

\usepackage{newfloat}
\usepackage{listings}
\DeclareCaptionStyle{ruled}{labelfont=normalfont,labelsep=colon,strut=off} 
\floatstyle{ruled}
\newfloat{listing}{tb}{lst}{}
\floatname{listing}{Listing}

\title{UniDoc: A Universal Large Multimodal Model for Simultaneous \par Text Detection, Recognition, Spotting and Understanding}

\author{
Hao Feng\textsuperscript{\rm 1,2,*},\quad Zijian Wang\textsuperscript{\rm 2,*},\quad Jingqun Tang\textsuperscript{\rm 2},\quad Jinghui Lu\textsuperscript{\rm 2},\\ Wengang Zhou\textsuperscript{\rm 1},\quad Houqiang Li\textsuperscript{\rm 1},\quad Can Huang\textsuperscript{\rm 2}\\
	haof@mail.ustc.edu.cn,\quad \{zhwg, lihq\}@ustc.edu.cn,\\ \{wangzijian.94, tangjingqun, lujinghui, can.huang\}@bytedance.com\\
}

\affiliations{
    \textsuperscript{\rm 1} University of Science and Technology of China \quad
    \textsuperscript{\rm 2} ByteDance \\
}

\begin{document}

\maketitle

\begin{figure*}[t]
	\centering
	\includegraphics[width=1.8\columnwidth]{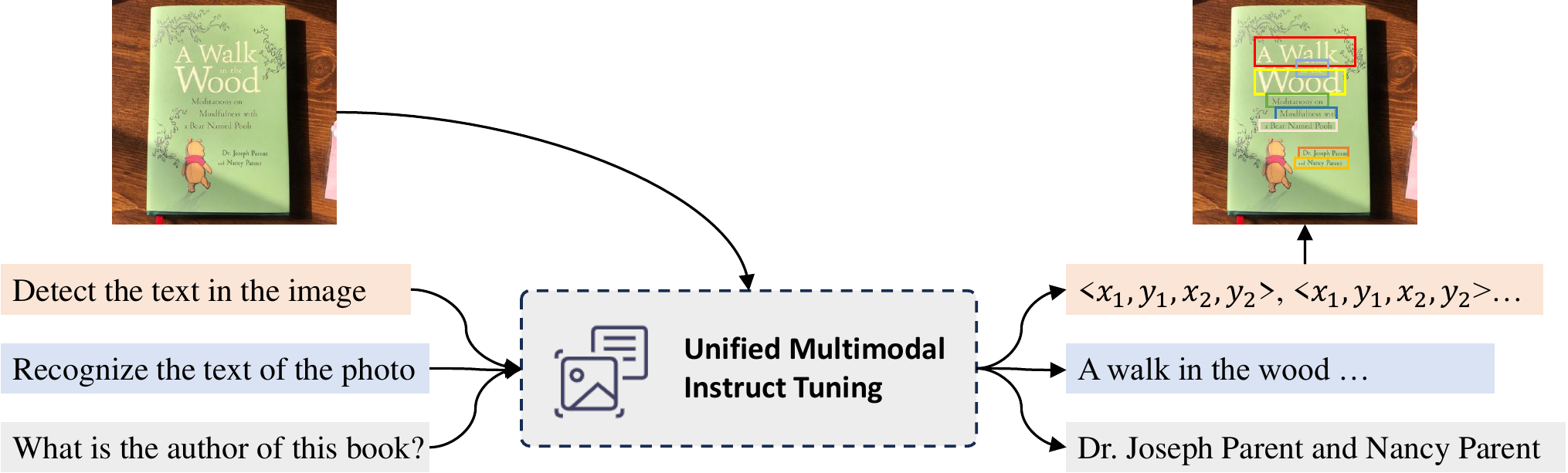}
	\caption{Framework of our UniDoc for simultaneous
		text detection, recognition, spotting (\emph{i.e.}, end-to-end text detection and recognition), and multimodal understanding in text-rich scenarios. It unifies these tasks into a single framework driven by natural language instructions. The unified multimodal instruct tuning not only broadens the repertoire of UniDoc, but also elevates the upper limit of its pre-existing capacities through the beneficial interactions among tasks.
	}
	\label{fig1}
\end{figure*}

\begin{abstract}
	In the era of Large Language Models (LLMs), 
	tremendous strides have been made in the field of multimodal understanding.
	However,
	existing advanced algorithms are limited to effectively utilizing the immense representation capabilities and rich world knowledge inherent to these large pre-trained models, and the beneficial connections among tasks within the context of text-rich scenarios have not been sufficiently explored.
	In this work, we introduce UniDoc, a novel multimodal model equipped with text detection and recognition capabilities, which are deficient in existing approaches. 
	Moreover, UniDoc capitalizes on the beneficial interactions among tasks to enhance the performance of each individual task.
	To implement UniDoc, we perform unified multimodal instruct tuning on the contributed large-scale instruction following datasets.
	Quantitative and qualitative experimental results show that UniDoc sets state-of-the-art scores across multiple challenging benchmarks.
	To the best of our knowledge, this is the first large multimodal model capable of simultaneous text detection, recognition, spotting, and understanding.
\end{abstract}

\section{Introduction}
Nowdays, considerable advancements have been observed in the domain of Large Language Models (LLMs), such as ChatGPT,~\footnote{https://openai.com/blog/chatgpt} BLOOM~\cite{scao2022bloom}, and LLaMA~\cite{touvron2023llama,touvron2023llama2}. These developments constitute significant strides towards the achievement of artificial general intelligence (AGI) and exhibit superior zero-shot proficiency across various linguistic applications. 
By employing these LLMs as language decoders, their Multimodal counterparts (LMMs), which include models like BLIP~\cite{li2023blip}, MiniGPT-4~\cite{zhu2023minigpt}, LLaVA~\cite{liu2023visual}, and mPLUG-Owl~\cite{ye2023mplug}, have showcased noteworthy efficacy in understanding visual and linguistic data.

While these large multimodal models exhibit astonishing zero-shot multimodal understanding capabilities, their comprehension of text-rich images remains limited~\cite{liu2023hidden}. 
To address this gap, LLaVAR~\cite{zhang2023LLaVAR} proposes incorporating a text recognition pre-training task to enhance the understanding of text-rich images. Besides, mPLUG-DocOwl~\cite{ye2023mplug} constructs a large-scale dataset about the document image understanding.
Although their text-rich scene understanding capabilities have shown notable promise, the vast potential of these pretrained large visual and language models remains largely unexplored and underutilized, analyzed next.

Firstly,
a salient absence of text detection capabilities is observed in the current large multimodal models.
Since these large visual and linguistic models are pre-trained on extremely large-scale datasets, they possess powerful representational capabilities and a wealth of world knowledge, endowing them with the ability to localize objects/text in images.
Their potential can be further harnessed and explored.
Secondly, the training strategies of advanced methods suffer from data distribution inconsistencies between the pre-training and fine-tuning phases~\cite{brown2020language}, leading to suboptimal performance.
Typically,
LLaVAR~\cite{zhang2023LLaVAR}
solely conducts text recognition tasks during the pre-training phase and proceeds with document understanding training in the fine-tuning phase.
Thirdly, 
text detection and recognition inherently fall under the umbrella of high-level scene understanding tasks, with the location and content of the text being associated with scene semantics.
Existing LMMs for text-rich image understanding have not effectively capitalized on these beneficial connections among OCR tasks~\cite{li2017towards} to enhance the performance on the individual tasks.

Formally,
we introduce UniDoc,
a universal large multimodal model for simultaneous
text detection, recognition, spotting, and understanding.
UniDoc aims to establish comprehensive OCR and multimodal understanding capabilities tailored for text-rich images.
We integrate all these tasks into a cohesive framework driven by natural language instructions for multimodal understanding, as shown in Fig.~\ref{fig1}.
Based on such a unified multimodal instruct tuning, not only have we endowed our UniDoc with various OCR capabilities, but the beneficial interactions among these tasks have also enhanced the performance across individual task.
To implement our UniDoc, we collected and annotated a large-scale instruction following dataset for this tasks.
Extensive quantitative and qualitative experimental results demonstrate the superior performance of UniDoc and its strong generalization ability.
To our best knowledge, this is the first large multimodal model capable of simultaneous text detection, recognition, spotting, and understanding.

In summary, we make three-fold contributions as follows:
\begin{itemize} 
	\item
	We introduce UniDoc, the first large multimodal model capable of simultaneous text detection, recognition, spotting, and multimodal understanding of text-rich images.
	\item
	We contribute a large-scale multimodal instruction tuning dataset, tailored for tasks of text detection, recognition, and spotting within text-rich images.
	\item
	We achieve state-of-the-art performance on multiple publicly available benchmark datasets. Moreover, we conduct extensive quantitative and qualitative experiments to validate the effectiveness of UniDoc.
\end{itemize}

\section{Related Work}
In this section,
we broadly review the recent research on instruction tuning and multimodal instruction tuning.

\begin{figure*}[t]
	\centering
	\includegraphics[width=2\columnwidth]{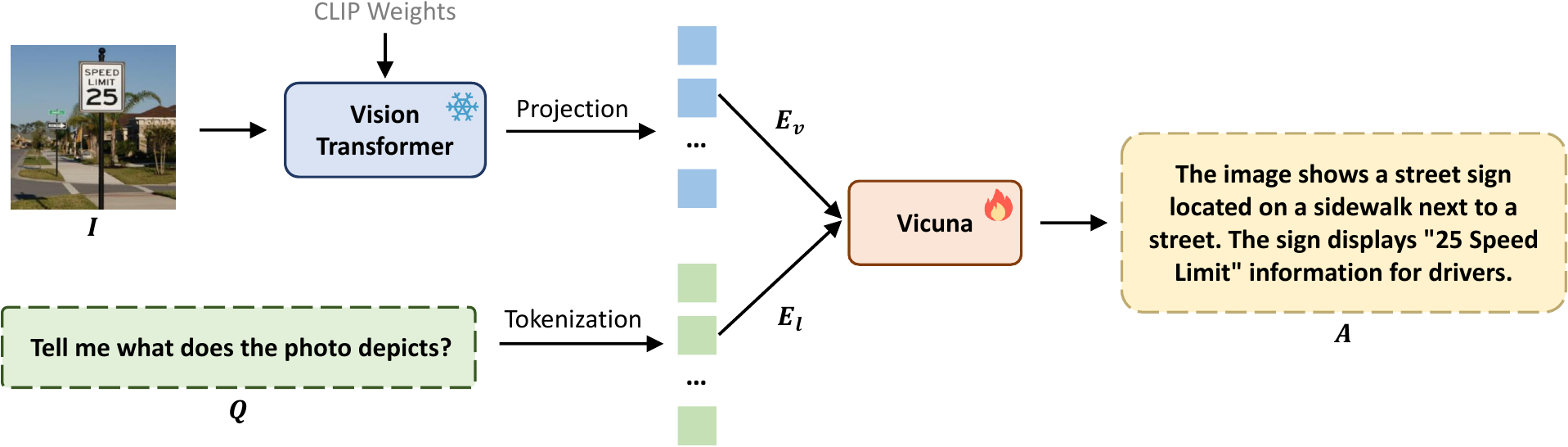}
	\caption{UniDoc Architecture. Given an image represented by $\bm{I}$ and a natural language instruction denoted as $\bm{Q}$, UniDoc synthesizes visual and textual cues. It extracts visual features from the input image $\bm{I}$ and assimilates textual cues originating from both the image $\bm{I}$ and the instruction $\bm{Q}$. Harnessing the extensive world knowledge embedded in the large language model (LLM), UniDoc performs coherent reasoning to produce contextually appropriate responses $\bm{A}$.}     
	\label{frame}
\end{figure*}

\subsection{Instruction Tuning}
Instruction tuning is an effective technique to align large language models (LLMs) with human intents.
It aims to teach language models to follow natural language (including prompt, positive or negative examples, and constraints etc.), to perform better multi-task learning on training tasks and generalization on unseen tasks.
Recently, models like GPT-3~\cite{brown2020language} and others have significantly leveraged instructional fine-tuning.
Typically, Stanford's Alpaca~\cite{alpaca} employs self-instruction~\cite{wang2022self} to provide a cost-effective approach to obtain instruction data for fine-tuning LLaMA. Vicuna~\cite{chiang2023vicuna} that is a instructional fine-tuned LLaMA based on dialogues between users and ChatGPT, achieves performance comparable to ChatGPT~\cite{zheng2023judging}.

\subsection{Multimodal Instruction Tuning}
Recent advancements in the confluence of natural language processing and computer vision have seen the rise of Large Multimodal Models (LMMs), which integrate large language models and visual encoders to address complex tasks involving both text and vision. Prominent works in this domain include MiniGPT-4~\cite{zhu2023minigpt}, which fuses components from BLIP-2~\cite{li2023blip} and Vicuna~\cite{chiang2023vicuna} for modality mapping and adopts a two-stage fine-tuning strategy. The LLaVA model, on the other hand, employs a supplementary linear layer to map visual features to the text space and undergoes additional fine-tuning under multimodal instructions. In the same vein, mPLUG-Owl from Alibaba's DAMO Academy incorporates Flamingo's Perceiver Resampler structure to facilitate visual and language modalities alignment. Another significant contribution is from InstructBLIP, which introduces a novel multimodal instruction dataset and uses Q-Former and Vicuna as an image encoder and language model respectively. Finally, X-LLM has introduced a Chinese multimodal instruction dataset and employs several adapters to map different modalities to the text space.
While these multimodal large models exhibit promising visual-linguistic understanding capabilities, their potential are yet to be fully harnessed in specific domains.

To bridge this divide, LLaVAR~\cite{zhang2023LLaVAR} puts forward the inclusion of a text recognition pre-training task, thus bolstering the comprehension of text-heavy imagery. In addition, mPLUG-DocOwl~\cite{ye2023mplug} has compiled an expansive dataset designed specifically for the fine-tuning of document comprehension tasks. Shikra~\cite{chen2023shikra} integrates LMMs with visual grounding ability by recasting detection task as a prompt-guided seq2seq task. 
Although these approaches somewhat augment the multimodal comprehension ability of models in text-rich scenarios, they fall short in offering a comprehensive ability for text detection, recognition and spotting. Moreover, they do not effectively harness the potential reciprocal enhancements that could be achieved by learning these capabilities in tandem.

\section{Methodology}

\subsection{Model Architecture}
Fig.~\ref{frame} presents an overview of our UniDoc. Our design follows the paradigm established by MiniGPT-4~\cite{zhu2023minigpt} and LLaVA~\cite{liu2023visual}. 

Specifically, given an input \emph{RGB} image $\bm{I} \in \mathbb{R}^{H\times W\times3}$ and a natural language instruction $\bm{Q}$, UniDoc first abstracts the visual features from $\bm{I}$ utilizing CLIP-ViT-L/14~\cite{radford2021learning} as the visual encoder. Both pre- and post- Transformer layer grid features are incorporated in our method. The extracted feature map is then flattened into a sequence of visual embedding sequence 
and projected into the embedding dimension of the LLM with a linear layer.
The output sequence $\bm{E}_v$ and then concatenated with embedding sequence $\bm{E}_l$ tokenized from the language instruction $\bm{Q}$. 

Thereafter, the concatenated embedding sequence are fed into Vicuna~\cite{chiang2023vicuna}, a large language model originating from the LLaMA~\cite{touvron2023llama} and specifically tuned with the instruction following data.
Vicuna~\cite{chiang2023vicuna} then generates the response based on the received visual and text cues.
Note that the visual embedding here can be considered as a soft prompt for LLM.

\subsection{Unified Multimodal Instruct Tuning}
Our training process is divided into two stages. 
Both stages employ our unified multimodal instruct tuning.
The first pre-training phase aims to align the output features from the pre-trained visual encoder with the feature space of the large language model. During the second fine-tuning stage, we further optimize the weights of the large language model.

Concretely, during the pre-training phase, we freeze both the pre-trained large visual and language models, training only the linear projector to align the visual and language features.
Our instruction following data involves four tasks: text detection, recognition, spotting, and image captioning. 
We argue that detection, recognition, and spotting inherently involve high-level semantic understanding, as the position and content of text within an image often have a strong correlation with their surrounding context.
The image captioning task enhances the model's understanding of natural scene images.
All of these tasks were performed in a natural language instruction following manner.

\setlength{\tabcolsep}{0.62mm}
\begin{table}[t]
	\small
	\centering
	\begin{tabular}{llcccc}
		\toprule
		\textbf{Satge}  & \textbf{Data}  & \textbf{Image} & \textbf{Instruction} & \textbf{\# Conv} & \textbf{Task} \\ 
		\midrule
		\multirow{2}{*}{Pre-train}  & LLaVA & CC3M       & CC3M          & 595K  & $\mathcal{C}$  \\
		& UniDoc & LAION        & OCR          & 600K  & $\mathcal{D},\mathcal{R},\mathcal{S},\mathcal{C}$      \\
		\midrule  
		\multirow{3}{*}{Fine-tune}  & LLaVA  & COCO    & GPT-4     & 158K    &   $\mathcal{U}$       \\
		& LLaVAR                & LAION                 & GPT-4                       & 16K    &    $\mathcal{D},\mathcal{R},\mathcal{S},\mathcal{U}$          \\ 
		& UniDoc                & LAION                 & GPT-4 + OCR                      & 186K    &    $\mathcal{D},\mathcal{R},\mathcal{S},\mathcal{U}$          \\ 
		\bottomrule
	\end{tabular}
	\caption{Summary of the dataset statistics. The symbols $\mathcal{C},\mathcal{D},\mathcal{R},\mathcal{S},\mathcal{U}$ correspond to the different instruction following tasks, namely, captioning, detection, recognition, spotting, and multimodal understanding.}
	
	\label{data_distri} 
\end{table}

\begin{table*}[t]
	\centering
	\setlength\tabcolsep{5pt}
	\resizebox{0.88\linewidth}{!}{%
		\begin{tabular}{lcccccccccccc}
			\hline
			\multirow{3}{*}{Method} & \multicolumn{3}{c}{Regular}  & \multicolumn{6}{c}{Irregular}                                         & \multicolumn{2}{c}{Occluded}  & \multirow{3}{*}{Avg.}\\ \cmidrule(lr){2-4} \cmidrule(lr){5-10} \cmidrule(lr){11-12} 
			& IIIT5K  & SVT     & IC13 & IC15 & SVTP       & CT80       & COCO & CTW     & TT & HOST                  & WOST               \\
			& 3000    & 647     & 857       & 1811       & 645        & 288        & 9896     & 1572    & 2201      & 2416                  & 2416   \\ \hline
			
			BLIP-2 OPT$\mathrm{_{6.7b}}$                & \textbf{\color{blue}{76.63}} & 80.22 & 82.96   & 69.35    & 73.33    & 76.04    & 48.68  & 61.70 & 63.52   & \textbf{\color{blue}{57.00}}     & \textbf{\color{blue}{68.00}}   &  68.22       \\
			BLIP-2 FlanT5$\mathrm{_{XXL}}$                & 76.60 & \textbf{\color{blue}{83.77}} & \textbf{\color{blue}{86.35}}   & \textbf{\color{blue}{70.84}}    & \textbf{\color{blue}{73.80}}   & 80.90    & 50.10  & 64.50 & 65.74   & \textbf{\color{red}{57.16}}               & \textbf{\color{red}{68.34}}      &  \textbf{\color{blue}{69.20}}           \\
			OpenFlamingo            & 68.20 & 74.19 & 74.10   & 63.61    & 73.49    & 67.71    & 45.52  & 53.94 & 57.84   & 48.18  & 60.55 &  60.04            \\
			MiniGPT-4                & 48.00 & 50.39 & 48.89   & 42.19    & 50.39    & 57.29    & 26.25  & 41.86 & 40.57   & 34.52  & 41.06   &42.68   \\
			mPLUG-Owl               & 74.43 & 77.74 & 82.15   & 65.21    & 72.71    & \textbf{\color{blue}{81.94}}    & \textbf{\color{blue}{50.42}}  & \textbf{\color{blue}{68.64}} & \textbf{\color{blue}{68.11}}   & 47.81 & 60.60    & 66.60          \\
			\textbf{UniDoc} & \textbf{\color{red}{90.60}} & \textbf{\color{red}{86.09}} & \textbf{\color{red}{87.51}} & \textbf{\color{red}{75.70}} & \textbf{\color{red}{77.05}} & \textbf{\color{red}{83.68}} & \textbf{\color{red}{61.51}} & \textbf{\color{red}{72.20}} & \textbf{\color{red}{76.24}}  & 48.92 & 62.96 & \textbf{\color{red}{72.40}}  \\
			\hline
			Supervised-SOTA         &  \textbf{96.63}    &  \textbf{93.04}      &  \textbf{96.73}      &  \textbf{85.70}       &  \textbf{89.30}       &  \textbf{89.93}       &  \textbf{64.42}     &  \textbf{78.57}    &  \textbf{80.13}      &  \textbf{73.10}   &  \textbf{81.58}          & \textbf{84.78}                  \\ 
			\hline
	\end{tabular}}
	\caption{Quantitative comparison with existing large multimodal models (LMMs) on text recognition benchmarks.
		Here the input prompt reads, ``Extract all the text in this photo".
		Performance metrics highlighted in red represent the highest achieved results, while those highlighted in blue denote the second-best performance.
	}
	\label{tab:text_reco}
\end{table*}

In the fine-tuning phase, we unfreeze both the large language model and the projector.
Besides the training tasks involved in the pre-training stage, we further incorporate an additional multimodal understanding task for text-rich images which requires a more advanced level of semantic comprehension.
The learning of these tasks mutually enhance each other.
Through this unified multi-modal unified instruction fine-tuning, UniDoc achieves a comprehensive recognition and understanding capability for text-rich scenarios.

\section{Dataset Construction}
To train the UniDoc,
we construct a large-scale multimodal instruction following dataset.
We detail it in the following.

\begin{figure}[t]
	\centering
	\includegraphics[width=1\columnwidth]{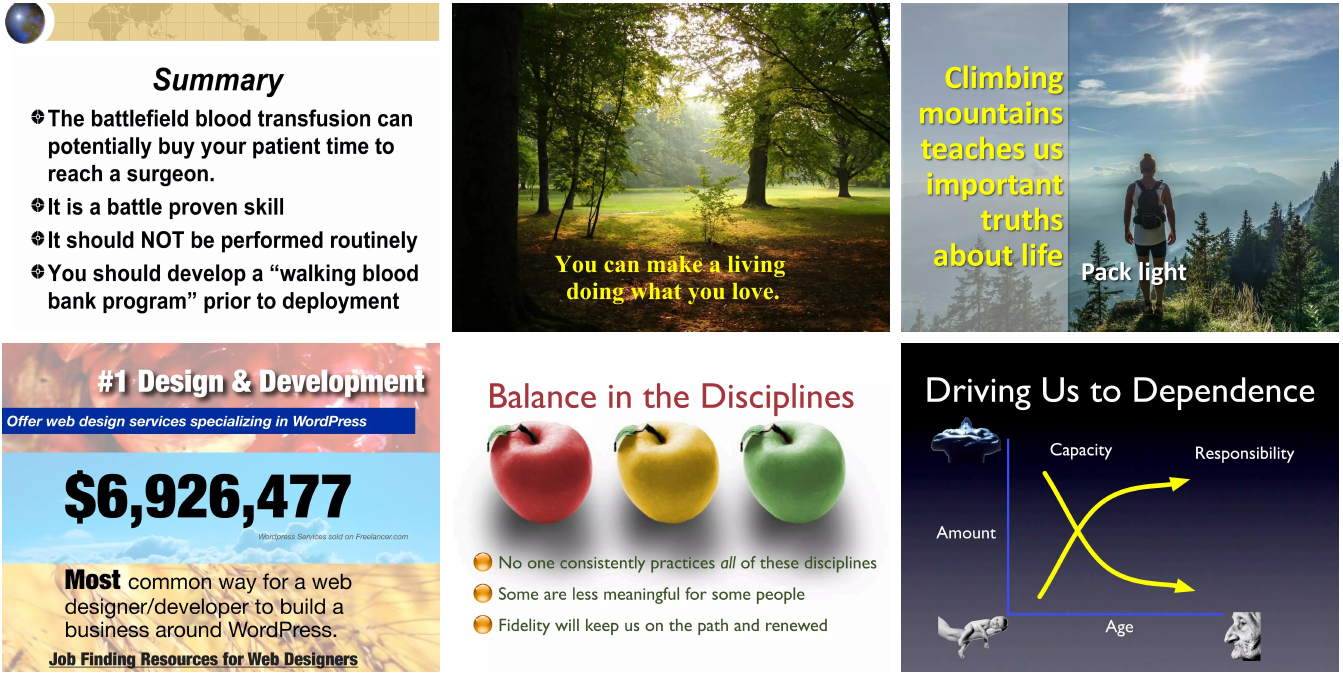}
	\vspace{-0.1in}
	\caption{Example instances from the proposed dataset, featuring diverse fonts in terms of size, style, and color, and a rich array of visual elements.}
	\label{dataset}
\end{figure}

\smallskip
\textbf{Pre-training.}
The pre-training data consists of two parts: one portion includes 595K natural scene images along with their captions, sourced from the CC3M dataset and filtered by LLaVA~\cite{liu2023visual}; the other portion comprises 600K image-text pairs from PowerPoint presentations that we created.
The data were collected from the ``Common Crawl" dataset, a vast web corpus containing publicly available web page.~\footnote{https://commoncrawl.org/}
We opt for PowerPoint files based on two primary considerations.
On one hand, PowerPoint presentations are characterized by a rich assortment of elements and their complex combinations, such as various fonts, images, tables, as shown in Fig.~\ref{dataset}. These elements are interrelated, making them highly conducive to training multimodal understanding tasks in text-rich scenarios.
On the other hand, the text within the slides is relatively large, making it legible for existing pre-trained visual models~\cite{radford2021learning}.
In other words, if the text in an image is too small, it becomes unrecognizable when input into the model. 

To ensure high-quality visuals suitable for our purposes, we conducted rigorous quality assurance checks, eliminating the noisy data to avoid any negative impact on training.
Specifically, we first applied text size optimization, excluding images with small-sized text.
Then, an in-house OCR tool accurately extracts the text and box annotations from each image and we constructed OCR instruction based on them.
The instructions here are categorized into three types: text detection, recognition, and understanding. Furthermore, we employed GPT-4 to generate diverse expressions for each type of instruction.
The data for detection, recognition, and spotting each account for one-third of the total.

\smallskip
\textbf{Fine-tuning.}
During fine-tuning,
we extend the 16K instruction following data collected from LAION-5B~\cite{schuhmann2022laion} and constructed by LLaVAR~\cite{zhang2023LLaVAR}.
Initially, we curated this dataset, employing the same cleansing methodology as used for the pre-training set. Subsequently, for each image, we constructed OCR instruction following data, adhering to the approach established during the pre-training phase.
The data for detection, recognition, and spotting each account for one-third of the total.
Furthermore, we further incorporated 150K OCR instruction data as the pre-training stage, in which detection, recognition, and spotting each constitute one-third of the total.

\begin{figure*}[t]
	\centering
	\includegraphics[width=1.9\columnwidth]{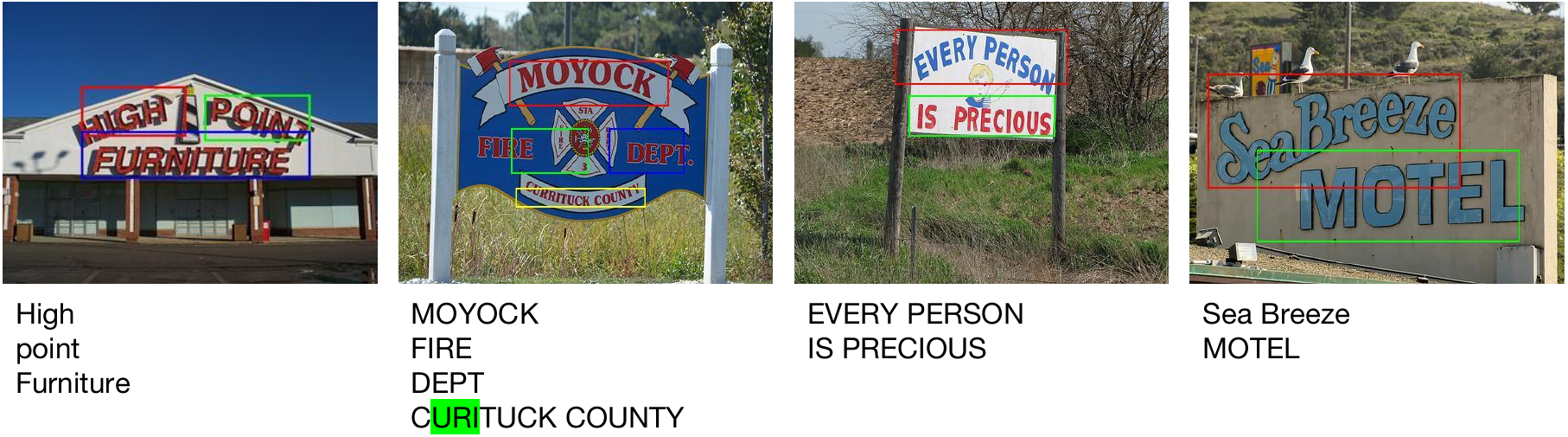}
	\caption{Visualization results of UniDoc for spotting on the CTW1500 dataset~\cite{shi2017detecting}. Our input prompt reads, ``Recognize all the text in this picture and return their positions [x1, y1, x2, y2]". From the structured response of UniDoc, we extracted the bounding boxes (visualized in the original image) and the corresponding recognized text (presented at the bottom). The green character highlight indicates incorrectly recognized text.}
	\label{fig_spotting}
\end{figure*}

\begin{figure*}[t]
	\centering
	\includegraphics[width=1.9\columnwidth]{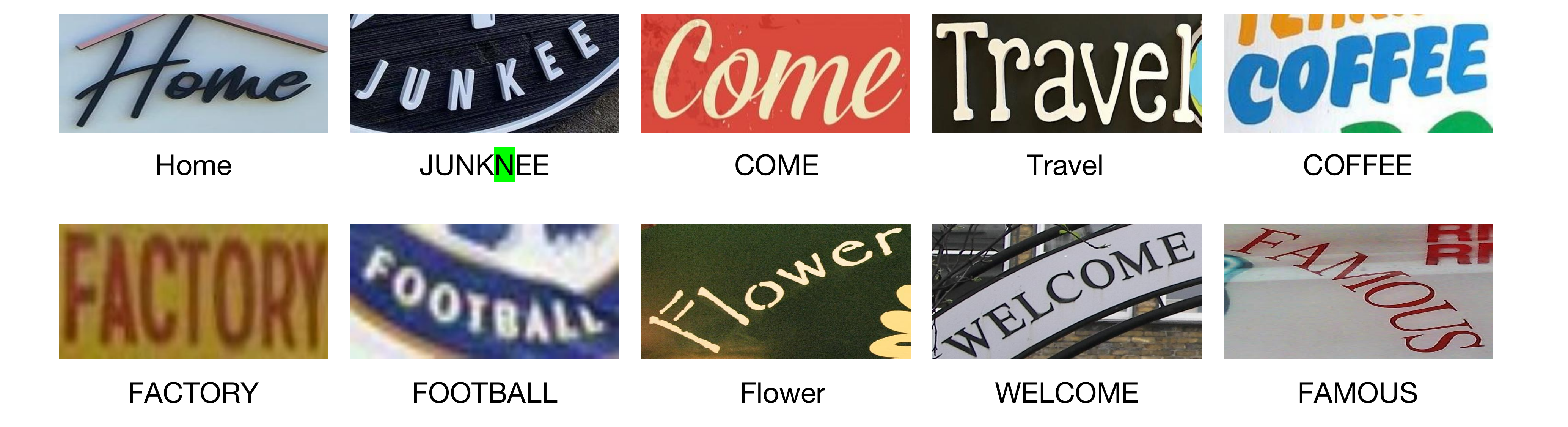}
	\caption{Visualization results of UniDoc for text recognition on the WordArt~\cite{xie2022toward} (visualized in the first and third row) and TotalText~\cite{ch2017total} (presented in the second and bottom row) dataset. Our input prompt reads, ``Extract all the text in this photo". The green character highlight indicates incorrectly recognized text.}
	\label{fig_recognize}
\end{figure*}

\section{Experiments}
\subsection{Training Details}
To implement UniDoc, we employed a one-cycle learning rate policy~\cite{smith2019super}. During the pre-training phase, the maximum learning rate was set to 1e-3, while for the fine-tuning phase, it was reduced to 1e-5. Moreover, the batch size was 128 for the pre-training and 
32 for the fine-tuning phase, respectively. The AdamW~\cite{loshchilov2017decoupled} optimizer was chosen for weight update. Both the pre-training and fine-tuning phases were executed employing eight A100 GPUs.
The pre-training and fine-tuning stages each consist of a single epoch.
In this paper, during both the training and inference phases, the input image resolution is consistently set at 224$\times$224. While larger input images are certain to yield better results due to the presence of more discernible text~\cite{zhang2023LLaVAR,ye2023mplug-doc}, that focus is beyond the scope of this study.

\setlength{\tabcolsep}{3mm}
\begin{table}[t]
	\small
	\centering
	\begin{tabular}{cccc} 
		\toprule
		\multirow{2}{*}{Method} & \multicolumn{3}{c}{Detection}  \\
		\cmidrule(rl){2-4}
		& CTW1500 & TotalText & TD500 \\
		\midrule
		\textbf{UniDoc} & 38.27 & 12.60 & 17.36 \\
		\bottomrule
	\end{tabular}
	\caption{Quantitative performance of UniDoc (F-score) on several scene text detection benchmark datasets. Here the input instruction is ``Output all the text locations in this photo".}
	\label{tab:det_results}
\end{table} 

\subsection{Evaluation Metrics}
We evaluate our UniDoc in a series of text-rich scenes from three perspectives (\emph{i.e.,} detection, recognition, and multimodal understanding).
For the task of text detection, we employed the F-score metric. For text recognition and visual question answering tasks, we adopted the accuracy metric, where a response generated by the model is considered correct if it contains the string present in the ground truth~\cite{liu2023hidden}.
In this paper, F-score and accuracy are respectively denoted as 
$\mathcal{F}$ and $\mathcal{A}$.

\begin{table*}[t]
	\centering
	\setlength\tabcolsep{1pt}
	\small
	\resizebox{0.9\linewidth}{!}{%
		\begin{tabular}{lcccccccccccc}
			\hline
			\multirow{3}{*}{Method} & \multicolumn{6}{c}{VQA}                                              & \multicolumn{3}{c}{KIE}        &\multicolumn{2}{c}{\multirow{3}{*}{HME100K}} &\multirow{3}{*}{Avg.}       \\ \cmidrule(lr){2-7}  \cmidrule(lr){8-10} 
			& STVQA   & OCRVQA  & TextVQA   & DocVQA  & InfoVQA & ChartQA   & FUNSD    & SROIE   & POIE \\
			& 5000    & 5000    & 5000      & 5349& 2801 & 1250       & 588      & 2503    & 6321      & \multicolumn{2}{c}{5000}                        \\ \hline
			BLIP-2 OPT$\mathrm{_{6.7b}}$                  & 13.36 & 10.58 & 21.18   &  0.82 & 8.82 & 7.44 & 0.00   & 0.00  & 0.02    & \multicolumn{2}{c}{0.00}                   & 5.86                  \\
			BLIP-2 FlanT5$\mathrm{_{XXL}}$                 & 21.70 & \textbf{\color{blue}{30.74}} & 32.18   &  4.86 & 10.17 &  7.20   & \textbf{\color{red}{1.19}}   & 0.20  & 2.52    & \multicolumn{2}{c}{0.04}                   & 11.02                     \\
			OpenFlamingo            & 19.32 & 27.82 & 29.08  & 5.05 & \textbf{\color{blue}{14.99}} & 9.12      & 0.85   & 0.12  & 2.12    & \multicolumn{2}{c}{0.00}                   & 11.07                      \\
			LLaVA                   & 22.08 & 11.36 & 28.86   &  4.49& 13.78 &   7.28      & \textbf{\color{blue}{1.02}}   & 0.12  & 2.09    & \multicolumn{2}{c}{0.04}                   & 9.03                      \\
			MiniGPT-4                & 14.02 & 11.52 & 18.72   & 2.97  & 13.32 &  4.32     & \textbf{\color{red}{1.19}}   & 0.04  & 1.31    & \multicolumn{2}{c}{0.00}                   & 6.54                       \\
			mPLUG-Owl               & 29.26 & 28.62 & \textbf{\color{blue}{40.28}}   &  \textbf{\color{red}{6.88}} & \textbf{\color{red}{16.46}} &  \textbf{\color{blue}{9.52}}      & \textbf{\color{blue}{1.02}}   & 0.64  & 3.26    & \multicolumn{2}{c}{\textbf{\color{blue}{0.18}}}                & \textbf{\color{blue}{13.74}}  \\  
			LLaVAR & \textbf{\color{blue}{30.36}} & 29.38 & 39.40 & \textbf{\color{blue}{6.73}}  & 12.25 & 8.00 & \textbf{\color{blue}{1.02}} & \textbf{\color{blue}{1.36}} & \textbf{\color{red}{6.48}} & \multicolumn{2}{c}{\textbf{\color{blue}{0.18}}} & 13.57 \\
			\textbf{UniDoc} & \textbf{\color{red}{30.78}} & \textbf{\color{red}{34.50}} & \textbf{\color{red}{40.72}} & 6.47 & 13.75 & \textbf{\color{red}{10.48}} & \textbf{\color{red}{1.19}} & \textbf{\color{red}{1.40}} & \textbf{\color{blue}{3.92}} & \multicolumn{2}{c}{\textbf{\color{red}{0.22}}} & \textbf{\color{red}{14.64}} \\
			\hline
			Supervised-SOTA         & \textbf{69.60}    & \textbf{68.10}    & \textbf{73.67}     &  \textbf{90.16}  &  	\textbf{36.82} &  \textbf{70.5}          & \textbf{93.12}    & \textbf{98.70}    & \textbf{79.54}     & \multicolumn{2}{c}{\textbf{64.29}}                    & \textbf{72.60}                             \\ \hline                        
	\end{tabular}}
	\caption{Quantitative comparison with existing large multimodal models (LMMs) on visual question answering (VQA), key information extraction (KIE), and handwritten mathematical expression (HMER) benchmarks.
		Performance metrics highlighted in red represent the highest achieved results, while those highlighted in blue denote the second-best performance.
	}
	\label{tab:text_reco_vqa_kie_res}
\end{table*}

\setlength{\tabcolsep}{1.2mm}
\begin{table}[t]
	\small
	\centering
	\begin{tabular}{ccccc} 
		\toprule
		\multicolumn{2}{c}{Training Task} & Detection & Recognition & Understanding   \\
		\cmidrule(rl){1-2}  \cmidrule(rl){3-3} \cmidrule(rl){4-4} \cmidrule(rl){5-5}
		Pre-train & Fine-tune & $\mathcal{F}$ & $\mathcal{A}$ & $\mathcal{A}$ \\
		\midrule
		& & 0.00 & 20.01 & 35.78 \\   
		\checkmark & & 0.00 & 84.13 & \textbf{41.28} \\
		& \checkmark & 27.89 & 88.93 & 40.46 \\
		\checkmark & \checkmark & \textbf{38.27} & \textbf{90.60} & 40.72 \\      
		\bottomrule
	\end{tabular}
	\caption{Ablation studies about the training tasking settings. The ``\checkmark" indicates that the corresponding training phase including the detection, recognition, and spotting task.}
	\label{tab:multitask}
\end{table}

\setlength{\tabcolsep}{0.2mm}
\begin{table}[t]
	\small
	\centering
	\begin{tabular}{ccccc} 
		\toprule
		\multirow{2}{*}{Experiment} & \multirow{2}{*}{Setting} & Detection & Recognition & Understanding  \\
		\cmidrule(rl){3-3} \cmidrule(rl){4-4} \cmidrule(rl){5-5}
		&  & $\mathcal{F}$ & $\mathcal{A}$ & $\mathcal{A}$ \\
		\midrule
		\multirow{2}{*}{index tokens}  & w/ & 31.28 & -  & - \\
		& w/o & \textbf{38.27} & -  & - \\  
		\midrule
		\multirow{2}{*}{instruction type}  & detection & 38.27 & - & - \\
		& spotting & \textbf{43.33} & - & - \\   
		\midrule
		\multirow{2}{*}{instruction type}  & recognition & - & 90.60 & -  \\
		& spotting & - & \textbf{91.30} & - \\ 
		\bottomrule
	\end{tabular}
	\caption{
		Ablation studies about variations in detection task configurations, and the impacts of the instruction type on text detection and recognition during inference.
	}
	\label{tab:otheraba}
\end{table}

\subsection{Comparison with Other LMMs}
We perform an exhaustive evaluation of publicly accessible large multimodal models (LMMs) and our UniDoc, assessing their efficacy across various benchmarks.
In the following, we compare and analyze the experimental results.

\smallskip
\textbf{Text Detection.}
Compared with the existing large multimodal models (LLMs), a unique capability of our UniDoc is its text detection ability. This stems from our approach of incorporating text detection as part of the unified multimodal instruction tuning.
In Table~\ref{tab:det_results}, we present the quantitative performance of our method on multiple scene text detection datasets, including CTW1500~\cite{shi2017detecting}, TotalText~\cite{ch2017total}, and TD500~\cite{yao2012detecting}.
Moreover,
as illustrated in Fig.~\ref{fig_spotting}, we provide examples showcasing UniDoc's text detection performance on the CTW1500 dataset~\cite{shi2017detecting}. 
It can be seen that the text is consistently detected in these images. 
Notably, 
the words in these images are located irregularly instead of in a straight horizontal line, and our training phase also does not involve the text detection tasks for such scene images. 
These findings validate our learning strategy and underscore the substantial generalization ability of LLMs.

\begin{figure}[t]
	\centering
	\includegraphics[width=0.82\columnwidth]{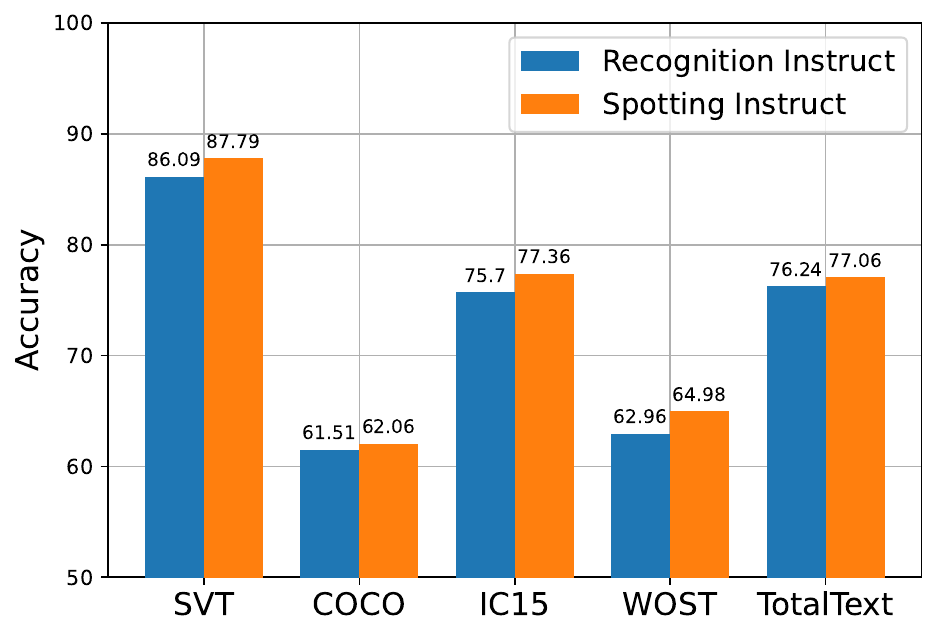}
	\label{fig0}
	\caption{Quantitative comparison on multiple recognition datasets based on the recognition instructions and spotting instructions. The $x$-axis represents the datasets. Spotting instruction consistently performs better.}
	\label{zhuzhuang}
	\vspace{-0.05in}
\end{figure}

\begin{figure}[t]
	\centering
	\includegraphics[width=1\columnwidth]{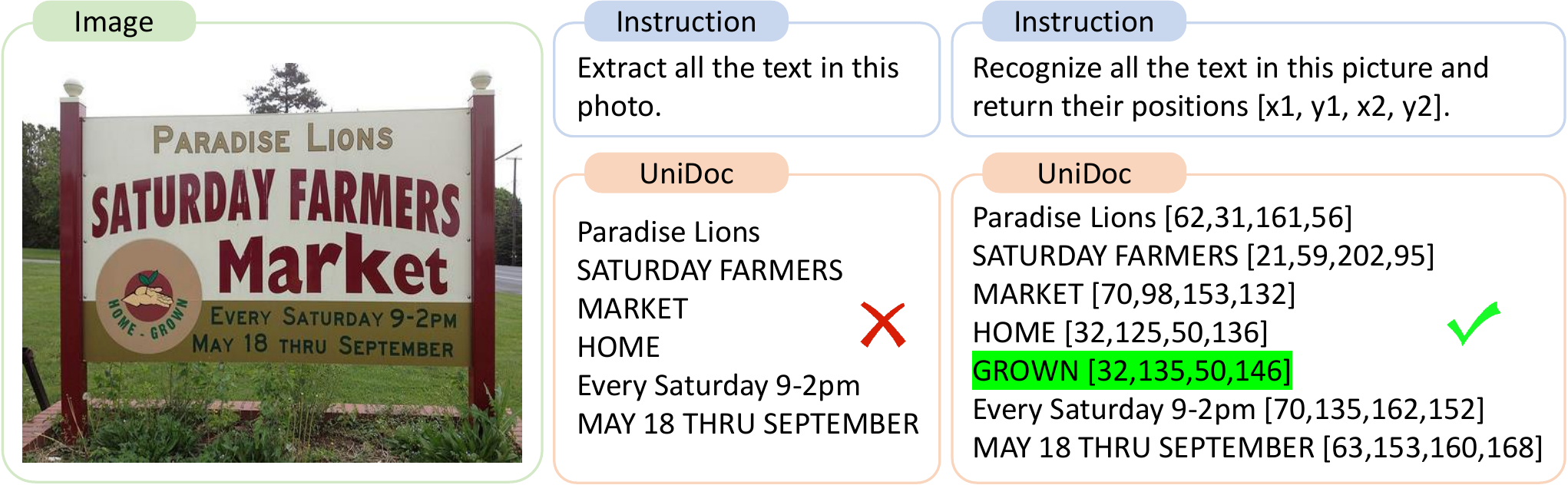}
	\vspace{-0.1in}
	\caption{A case study illustrating the impact of detection (left) and spotting (right) instructions on the response. Spotting effectively mitigates recognition omissions.
	}
	\label{instruct_type_case}
\end{figure}

\smallskip
\textbf{Text Recognition.}
Furthermore, we extend our evaluation to assess the text recognition capacity of UniDoc. 
To commence, as shown in Table~\ref{tab:text_reco}, UniDoc achieves a series of state-of-the-art scores across numerous benchmark datasets for text recognition.
It is noteworthy that these datasets encompass a diverse array of text-rich images, including document text, artistic text, handwritten text, scene text, and more. 
Moreover, as depicted in Fig.~\ref{fig_spotting} and Fig.~\ref{fig_recognize}, we showcase recognition results of UniDoc on CTW1500~\cite{shi2017detecting}, WordArt~\cite{xie2022toward} and TotalText~\cite{ch2017total} dataset.
Although these images involves varying fonts, styles, image blurriness, and non-horizontal text distributions, our UniDoc consistently manifests a remarkable ability to accurately recognize the embedded text within them.

\begin{figure*}[t]
	\centering
	\includegraphics[width=1.96\columnwidth]{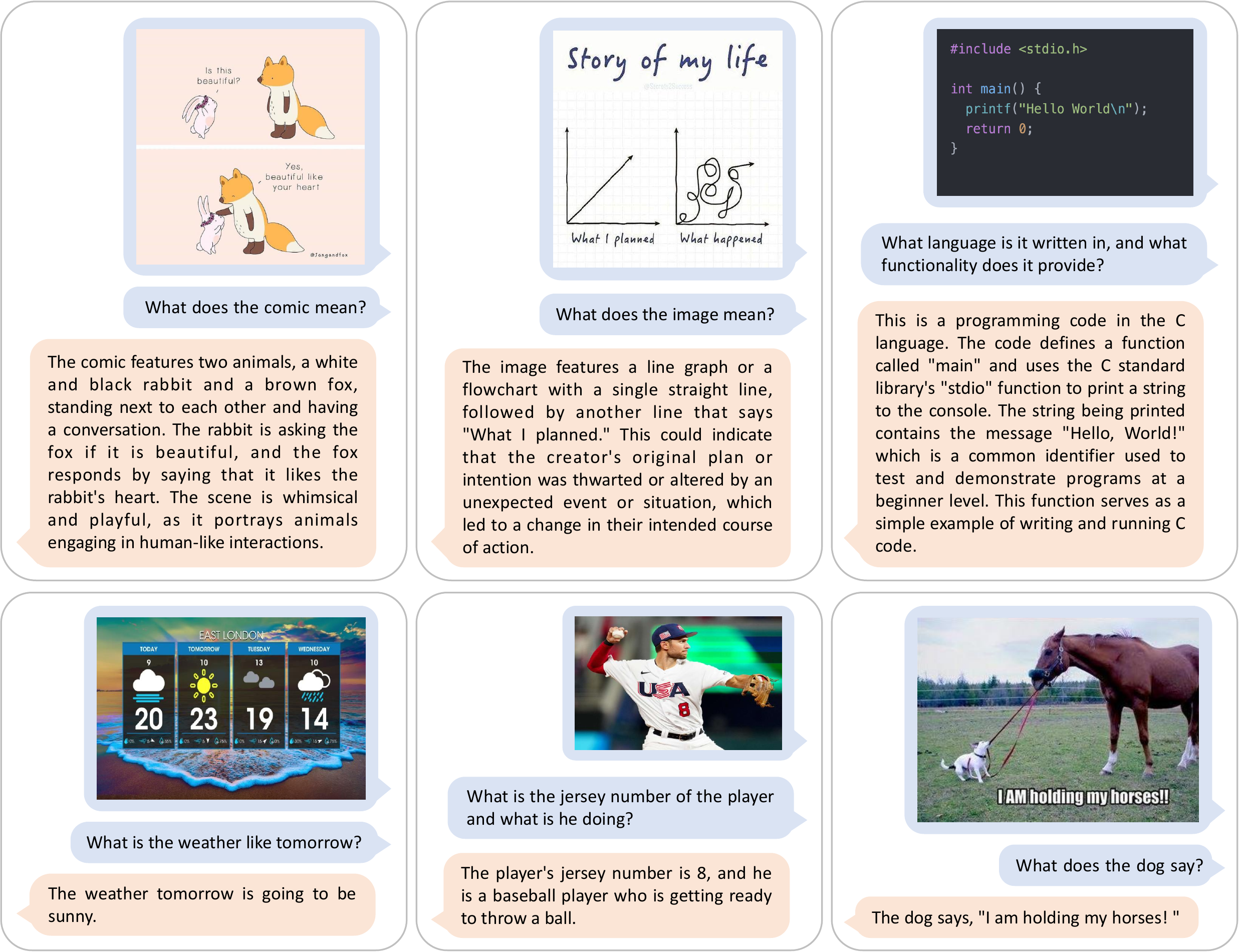}
	\caption{Visualization results of UniDoc for visual question answering (VQA) in text-rich scenes. The involved images exhibit diverse visual elements, and the text within them varies in font, size, color, and style.}
	\label{fig_understanding}
	\vspace{-0.08in}
\end{figure*}

\smallskip
\textbf{Multimodal Understanding.}
We conduct both quantitative and qualitative assessments of UniDoc's multimodal understanding performance. Specifically, as presented in Table~\ref{tab:text_reco_vqa_kie_res}, UniDoc achieves state-of-the-art and comparable performance on several benchmark datasets. Besides, as illustrated in the Fig.~\ref{fig_understanding}, we provide examples of multimodal question-answering focused on text-based scenarios. 
It can be seen that
UniDoc effectively integrates the visual cues from the input image and the textual cues from both the image and instructions. Leveraging the inherent world knowledge of the large language model (LLM), it then engages in coherent reasoning to generate corresponding responses.

\subsection{Ablation Studies}
In this section, we conduct ablation studies to validate the efficacy of core settings and components in our UniDoc.
In all experiments, for the tasks of text detection, recognition, and multimodal understanding, we report the performance on the CTW1500~\cite{shi2017detecting}, IIIT5K~\cite{mishra2012scene}, and TextVQA~\cite{singh2019towards} benchamrk datasets, respectively.

\smallskip
\textbf{Impact of Unified Multimodal Instruct Tuning.}
During the pre-training phase, the instruction-following data we trained encompasses text detection, recognition, and spotting tasks. In the fine-tuning phase, the instruction-following data was further augmented with tasks concerning multimodal understanding. 
we investigate the impact of learning these tasks (\i.e., text detection, recognition, and spotting) on the final performance.
As illustrated in Table~\ref{tab:multitask}, incorporating the learning of them in individual phases led to enhancements not only in detection and recognition performance, but also in multimodal understanding.
Furthermore, incorporating these tasks in both stages yielded the best performance.
These results demonstrate that there exists a beneficial interplay and synergy among these tasks. 
We argue that such a multi-task learning strategy not only endows Large Multimodal Models (LMMs) with comprehensive capabilities, but also bolsters their inherent abilities.

\smallskip
\textbf{Impact of the Formulation of the Detection Task.}
In our default setting, we directly predict the integer coordinates of the text region bounding boxes. Given that our input images are all of the size 224$\times$224, these coordinates are normalized to the range [0, 223]. An alternative approach is to set up an additional 224 tokens to represent both the horizontal and vertical coordinates in the range [0, 223]~\cite{chen2021pix2seq}. As shown in Table~\ref{tab:otheraba}, in terms of text detection capabilities, the introduction of additional positional index tokens did not yield a performance gain.

\smallskip
\textbf{Impact of Instruction Template Type.}
In our UniDoc, the detection results can originate from either the detection or the spotting instructions. Similarly, our recognition outcomes can be sourced from either the recognition or the spotting instructions. Consequently, we evaluate the impact of using different types of instructions on the performance of detection and recognition.
As shown in Table~\ref{tab:otheraba}, the text detection and recognition performance based on the spotting instruction works better.
This is likely because in autoregressive generation, spotting instruction template makes model provide explicit location information in its responses, enhancing the recognition performance. The same applies to detection tasks.
The two tasks are mutually complementary.
In Fig.~\ref{zhuzhuang}, we perform quantitative comparisons on a broader range of recognition benchmarks.
Besides, as shown in Fig.~\ref{instruct_type_case},
we further provide a case to illustrate this finding.

\section{Conclusion}
In this work,
we introduce UniDoc,
a universal large multimodal model for simultaneous text detection, recognition, spotting, and understanding.
Through our proposed unified multimodal instruct tuning,
UniDoc effectively leverages the beneficial interactions among text-based tasks, not only addressing the shortcomings of existing large multimodal models, but also enhancing their original capabilities.
To implement UniDoc,
we contribute a large-scale multimodal instruction following dataset.
Experiments show that our UniDoc sets state-of-the-art scores across multiple benchmarks.
Besides,
we perform extensive studies to validate its effectiveness.
Currently, UniDoc is unable to extract fine-grained visual features for detection and recognition, and the resolution of input images remains a limitation. In the future, we will consider addressing these issues.

\bibliography{egbib}

\end{document}


\maketitle

\begin{table}[t]
	\centering
	\begin{tabular}{|c|m{6.5cm}|}
		\hline
		\textbf{No.} & \multicolumn{1}{c|}{\textbf{Instruction}} \\
		\hline
		1 &
		Output all the text's locations in \textless term\textgreater. \\
		\hline
	\end{tabular}
	\caption{Instruction template for the text detection task. In this template, the placeholder \textless term\textgreater\ is used to represent descriptors of image types, such as ``image" or ``picture," etc., thereby enhancing the diversity of the templates.}
	\label{tab:det_instru}
\end{table}

\begin{table}[t]
	\centering
	\begin{tabular}{|c|m{6.5cm}|}
		\hline
		\textbf{No.} & \multicolumn{1}{c|}{\textbf{Instruction}} \\
		\hline
		1 & Identify the text in \textless term\textgreater. \\
		\hline
		2 & Identify the text information in \textless term\textgreater. \\
		\hline
		3 & Detect all the text in \textless term\textgreater. \\
		\hline
		4 & What's the text in \textless term\textgreater? \\
		\hline
		5 & Extract all the text in \textless term\textgreater. \\
		\hline
		6 & Identify the text in \textless term\textgreater. \\
		\hline
		7 & Recognize all the text in \textless term\textgreater. \\
		\hline
		8 & Detect and recognize the text in \textless term\textgreater. \\
		\hline
		9 & Find all the text in \textless term\textgreater. \\
		\hline
		10 & Parse all the text in \textless term\textgreater. \\
		\hline
	\end{tabular}
	\caption{Instruction templates for the text recognition task. In these templates, the placeholder \textless term\textgreater\ is used to represent descriptors of image types, such as ``image" or ``picture," etc., thereby enhancing the diversity of the templates.}
	\label{tab:reco_instru}
\end{table}

\begin{figure}[t]
	\centering
	\includegraphics[width=1\columnwidth]{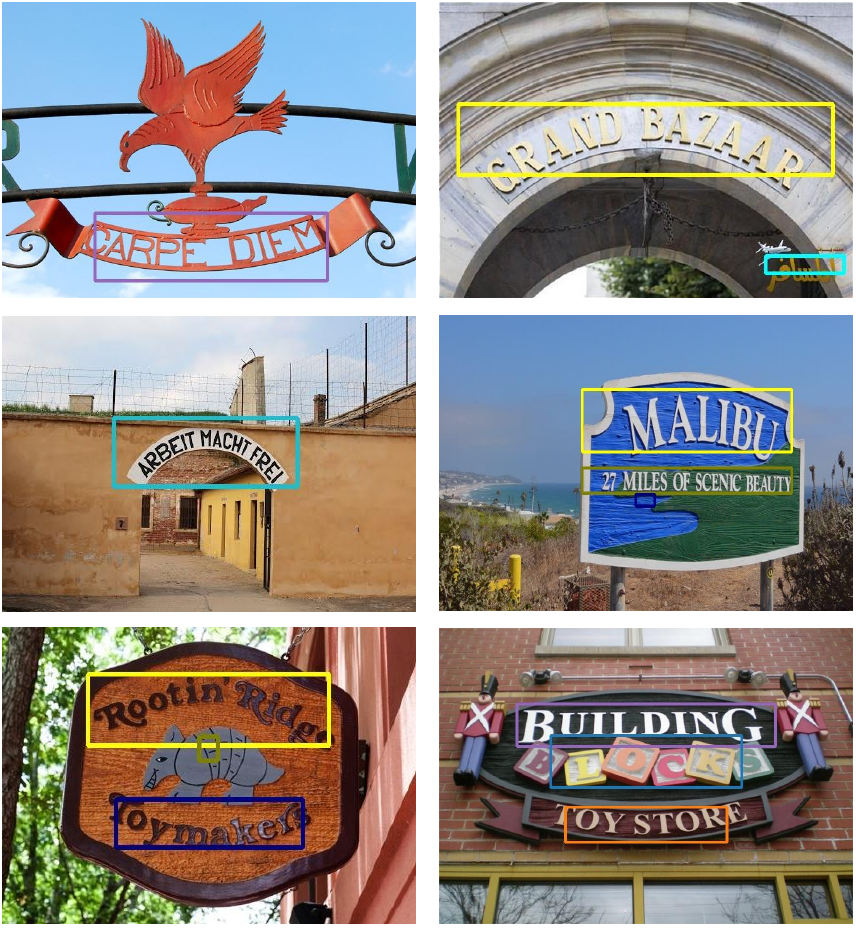}
	\caption{Visualization results of UniDoc for detection on the TotalText dataset~\cite{ch2017total}. Our input prompt reads, ``Output all the text's locations in the photo.". From the structured response of UniDoc, we extracted the bounding boxes (visualized in the original image).}
	\label{fig:supp_det_tt}
\end{figure}

\begin{table*}[t]
	\centering
	\begin{tabular}{|c|p{7cm}|c|p{7cm}|}
		\hline
		\textbf{No.} & \multicolumn{1}{c|}{\textbf{Instruction}} & \textbf{No.} & \multicolumn{1}{c|}{\textbf{Instruction}} \\
		\hline
		1 & Identify the text in \textless term\textgreater\ and return the positions. & 6 & Recognize all the text in \textless term\textgreater\ and return their positions [x1, y1, x2, y2]. \\
		\hline
		2 & Extract all the text in \textless term\textgreater\ and return their coordinates. & 7 & Identify the text in \textless term\textgreater\ and return their [x1, y1, x2, y2] coordinates. \\
		\hline
		3 & Identify the text in \textless term\textgreater\ and return their positions in the format of [x1, y1, x2, y2]. & 8 & Recognize all the text in \textless term\textgreater\ return their coordinates in the format of [x1, y1, x2, y2]. \\
		\hline
		4 & Detect the text in \textless term\textgreater\ and return their coordinates in the format of [x1, y1, x2, y2]. & 9 & Find all the text in \textless term\textgreater\ and return their coordinates in the format of [x1, y1, x2, y2]. \\
		\hline
		5 & Find all the text in \textless term\textgreater\ and return their positions represented in the format of [x1, y1, x2, y2]. & 10 & Parse all the text in \textless term\textgreater\ and return their coordinates in the format of [x1, y1, x2, y2]. \\
		\hline
	\end{tabular}
	\caption{Instruction templates for the text spotting task. In these templates, the placeholder \textless term\textgreater\ is used to represent descriptors of image types, such as ``image" or ``picture," etc., thereby enhancing the diversity of the templates.}
	\label{tab:spot_instr}
\end{table*}

\begin{figure*}[t]
	\centering
	\includegraphics[width=1.9\columnwidth]{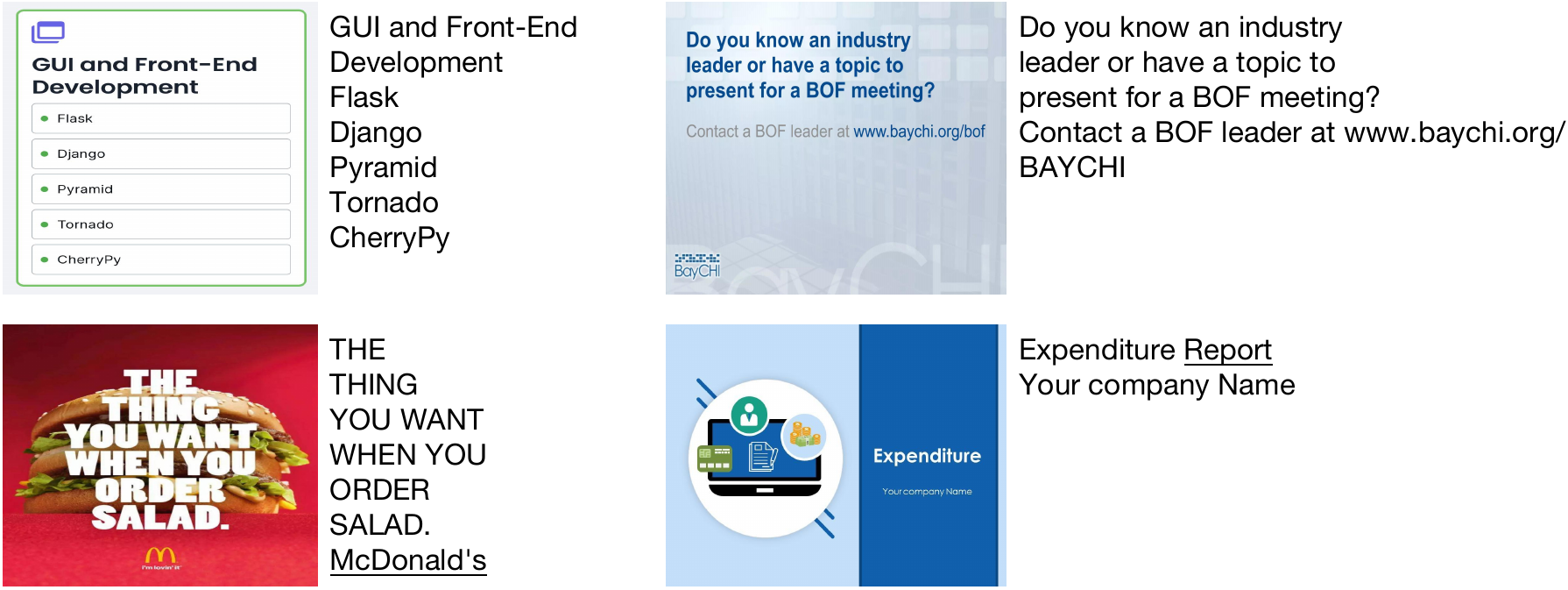}
	\caption{Visualization results of UniDoc for recognition on some digitized text-rich images. Our prompt reads, ``Recognize all the text in this picture and return their positions [x1, y1, x2, y2]". From the structured response, we extracted the recognized text. The underlined text here indicates some instances of erroneous recognition.}
	\label{fig:supp_reco_doc}
\end{figure*}

\begin{figure}[t]
	\centering
	\includegraphics[width=1\columnwidth]{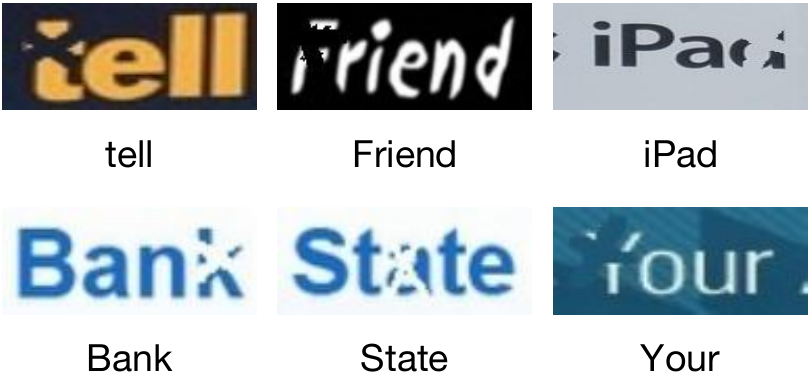}
	\caption{Visualization results of UniDoc for recognition of the line-level text images on the Host dataset~\cite{wang2021two}. We prompted UniDoc with the input, "Identify the text in this picture." From the structured response of our UniDoc, we extracted the recognized text displayed below each image. In this dataset, the text within some of the images may be partially missing or incomplete.}
	\label{fig:supp_reco_host}
\end{figure}

\begin{figure*}[t]
	\centering
	\includegraphics[width=2\columnwidth]{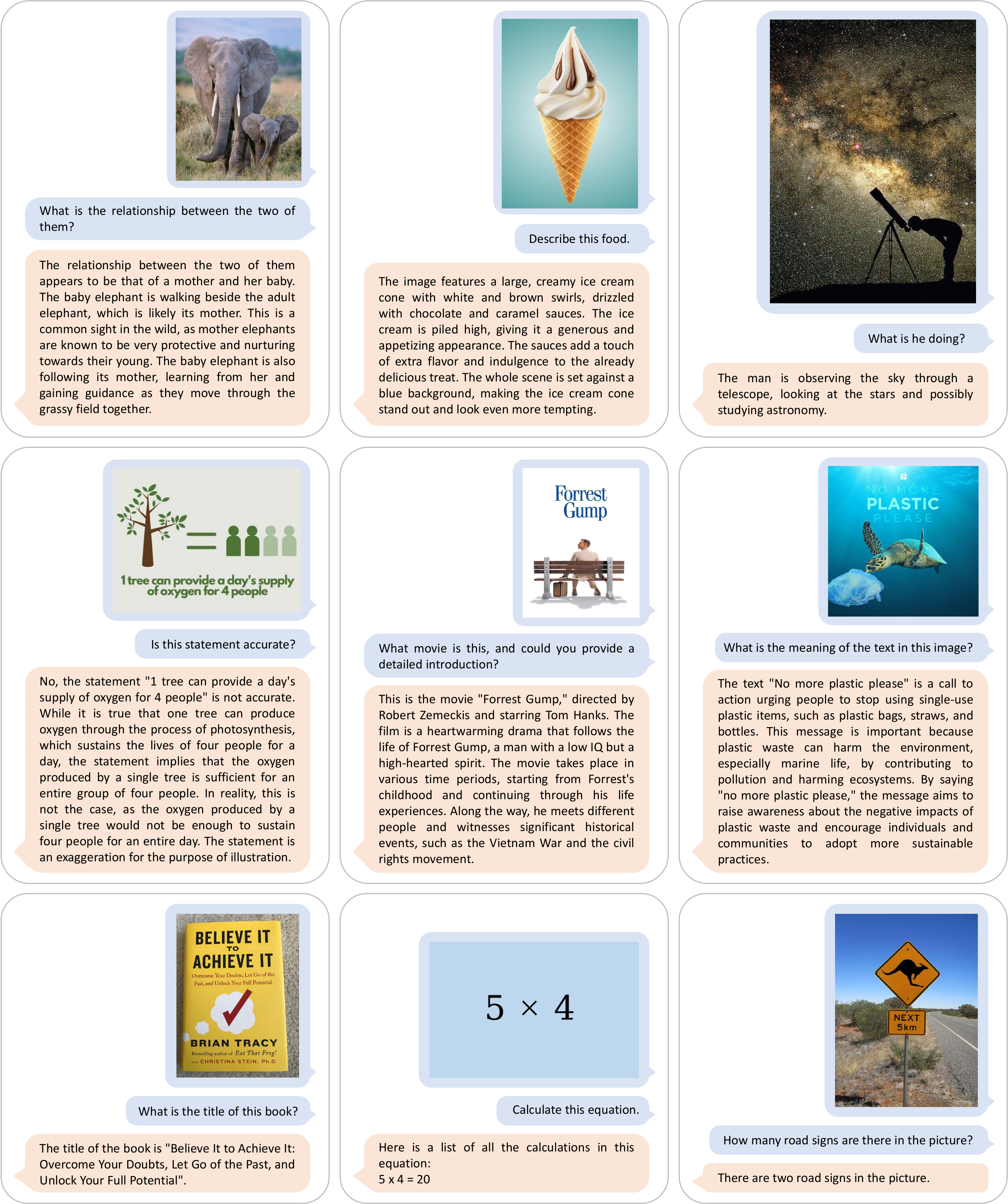}
	\caption{Visualization results of UniDoc for multimodal understanding across diverse image scenarios. UniDoc effectively extracts visual and textual information from images, content of instructions, and leverages the extensive world knowledge of large language models for rational inference and responses.}
	\label{fig:supp_undertand_case}
\end{figure*}

\section*{A. Prompt Template}
As shown in Table~\ref{tab:det_instru},~\ref{tab:reco_instru}, and~\ref{tab:spot_instr}, we present the instruction templates in our instruction following data for three distinct tasks, $i.e.$, text detection, recognition, and spotting.
These instruction templates were generated based on GPT-4, and we ensured the diversity of these instructions by varying the action verbs, descriptors of the images, and employing the question formats.
Note that in Table~\ref{tab:det_instru} there is only one template for our text detection task, which is designed to prevent confusion with the recognition instructions and to avoid adversely affecting network training.

\section*{B. Additional Qualitative Results}
In this section, we will demonstrate examples of text detection, recognition, and multimodal understanding by our UniDoc on several other benchmark datasets.

\smallskip
\textbf{Text Detection.}
As illustrated in Fig.~\ref{fig:supp_det_tt}, we present the text detection performance of UniDoc on the TotalText test set~\cite{ch2017total}.
We can see that the text in the images can be reasonably well detected.

\smallskip
\textbf{Text Recognition.}
We further showcase the ability of UniDoc to perform recognition on some electronic text-rich images,
unlike the natural scene images presented in the manuscript.
The results are presented in Fig.~\ref{fig:supp_reco_doc}.
We can observe that most of the text in such images can be recognized, yet our UniDoc sometimes tends to introduce some of its own conjectures when performing recognition.

In Fig.~\ref{fig:supp_reco_host}, we showcase examples of text recognition by UniDoc on line-level Host dataset~\cite{wang2021two}. In this dataset, the text within the images of this dataset may be missing or incomplete. However, our UniDoc is still able to accurately recognize the text.

\smallskip
\textbf{Multimodal Understanding.}
As shown in Fig.~\ref{fig:supp_undertand_case}, we further provide examples of UniDoc's capabilities in multimodal understanding across diverse image scenarios. We can observe that UniDoc can effectively incorporate the visual and textual information from images, the context of instructions, and leverages the rich world knowledge of large language models to generate reasonable responses.

\bibliography{egbib}